\documentclass[conference]{IEEEtran}
\IEEEoverridecommandlockouts

\usepackage{cite}
\usepackage{amsmath,amssymb,amsfonts}
\usepackage{algorithmic}
\usepackage{graphicx}
\usepackage{textcomp}
\usepackage{xcolor}

\usepackage{orcidlink}
\usepackage[acronym,toc]{glossaries}
\newacronym{cnn}{CNN}{Convolutional Neural Network}
\newacronym{dl}{DL}{Deep Learning}
\newacronym{ai}{AI}{Artificial Intelligence}
\newacronym{dnn}{DNN}{Deep Neural Networks}
\newacronym{ml}{ML}{Machine Learning}
\newacronym{gan}{GAN}{Generative Adversarial Network}
\newacronym{cv}{CV}{Confidence Value}
\newacronym{de}{DE}{Differential Evolution}
\newacronym{opencv}{OpenCV}{Open Computer Vision}
\newacronym{cpu}{CPU}{Central Processing Unit}
\newacronym{gpu}{GPU}{Graphic Processing Unit}
\newacronym{cuda}{CUDA}{Compute Unified Device Architecture}
\newacronym{ram}{RAM}{Random Access Memory}
\newacronym{dfdc}{DFDC}{Deepfake Detection Challenge}
\newacronym{roc}{ROC}{receiver operating characteristics}
\makeglossaries

\def\BibTeX{{\rm B\kern-.05em{\sc i\kern-.025em b}\kern-.08em
    T\kern-.1667em\lower.7ex\hbox{E}\kern-.125emX}}
    
\begin{document}

\title{Mitigating Adversarial Attacks in Deepfake Detection: An Exploration of Perturbation and AI Techniques}

\author{\IEEEauthorblockN{Saminder Dhesi, Laura Fontes\orcidlink{0000-0003-0171-7436}, Pedro Machado\orcidlink{0000-0003-1760-3871}, Isibor Kennedy Ihianle\orcidlink{0000-0001-7445-8573}, David Ada Adama\orcidlink{0000-0002-2650-857X}}
\IEEEauthorblockA{\textit{Department of Computer Science},  \\ Nottingham Trent University\\
Nottingham, UK\\
\{N0774151,N1119003\}@my.ntu.ac.uk\\
\{pedro.machado,isibor.ihianle,david.adama\}@ntu.ac.uk}
}

\maketitle

\begin{abstract}
Deep learning constitutes a pivotal component within the realm of machine learning, offering remarkable capabilities in tasks ranging from image recognition to natural language processing. However, this very strength also renders deep learning models susceptible to adversarial examples, a phenomenon pervasive across a diverse array of applications. These adversarial examples are characterized by subtle perturbations artfully injected into clean images or videos, thereby causing deep learning algorithms to misclassify or produce erroneous outputs. This susceptibility extends beyond the confines of digital domains, as adversarial examples can also be strategically designed to target human cognition, leading to the creation of deceptive media, such as deepfakes. Deepfakes, in particular, have emerged as a potent tool to manipulate public opinion and tarnish the reputations of public figures, underscoring the urgent need to address the security and ethical implications associated with adversarial examples. 
This article delves into the multifaceted world of adversarial examples, elucidating the underlying principles behind their capacity to deceive deep learning algorithms. We explore the various manifestations of this phenomenon, from their insidious role in compromising model reliability to their impact in shaping the contemporary landscape of disinformation and misinformation. To illustrate progress in combating adversarial examples, we showcase the development of a tailored Convolutional Neural Network (CNN) designed explicitly to detect deepfakes, a pivotal step towards enhancing model robustness in the face of adversarial threats. Impressively, this custom CNN has achieved a precision rate of 76.2\% on the DFDC dataset.
\end{abstract}


\section{Introduction} \label{sec:intro}
Nowadays, the falsification of images and videos on digital platforms is not a new problem, but the prior techniques for detecting these were mainly focused on face-swapping visualisation and its impact on the reputation of celebrities. Deepfakes, a type of synthetic media generated through \gls*{dl} and \gls*{ml} methods, were first introduced in 1997. To detect these falsified media, classification techniques are required that focus on facial recognition through different features such as the eyes, lips, or cheeks. The most commonly used approach is binary image classification with supervised learning, which distinguishes between real and fake. However, recent advancements have led to the development of more effective deepfake detection systems using \glspl*{gan} or Autoencoders. But these require large datasets, and the limited availability of image and video data can be a challenge to the training models which results in poor performance, reduced predictiveness and responsiveness.

To overcome the challenges posed by deepfakes, a pairwise learning approach can be employed to evaluate the similarity scores between inputs. Deepfakes involve the forging of faces, and the detection and recognition of real and fake faces is a major challenge due to factors such as lighting and facial expressions. Deep learning methods are prioritised for extracting discriminative features to improve robustness.

Adversarial examples in deepfake detection present a challenge to model accuracy, as these perturbed inputs can exploit vulnerabilities and manipulate the model output. To counteract these attacks, techniques such as cropping and compression can be used, as well as evaluation of the robustness of sub-sampled networks. The deepfake detection model architecture can be designed to withstand either white-box or black-box attacks, using face recognition and \gls*{cnn} for image classification and considering different perturbations in detection methods.

Differential privacy is used to defend against adversarial attacks on \gls*{ml} models by balancing accuracy and information impact within the dataset. To address potential malicious use of \gls*{ml} systems, four main categories of mitigating strategies have been proposed: rate limiting, regression testing, anomaly detection, and human intervention. The countermeasures involve reducing the impact of perturbations through data encryption, data sanitation, and robust statistics, in order to minimise errors during training and testing.

In this article, we examine the significance of deepfake detection by examining the presence of real images and the robustness of the data. The article will delve into the optimisation of Deep Learning models, such as \glspl*{cnn} and \glspl*{gan}, for deepfake detection. The models will operate on a frame-by-frame basis to analyse the temporal dependencies in videos, with Neural Networks being utilised to accurately classify fake videos. The main focus of this study is to address the challenges posed by deepfake generation and the use of publicly available, standardized datasets, such as the 140k real vs. fake and deepfake detection challenge available on Kaggle\footnote{Available online, \protect\url{https://www.kaggle.com/}, last accessed 09/02/2023}. The structure of the article is as follows: related work is analysed in Section \ref{sec:lr}, the methodology is presented in Section \ref{sec:method}, the results analysis in Section \ref{sec:results} and the conclusions and future work in Section \ref{sec:conclusions}.

\section{Related work}\label{sec:lr}
Adversarial attacks are a growing concern in the field of artificial intelligence and machine learning. These attacks involve manipulating input data in a way that causes a model to make incorrect predictions, often with the intent of compromising security or privacy. Some of the key challenges posed by adversarial attacks include the need to carefully design models that are robust against such manipulation, to develop methods for detecting and defending against attacks in real-world scenarios, and to improve understanding of the underlying reasons for why models can be vulnerable to these types of attacks \cite{ilahi2021}. In response to these challenges, researchers and practitioners are exploring a variety of countermeasures to improve the robustness of machine learning models. These countermeasures range from modifying model architecture and training procedures, to developing new techniques for detecting and defending against adversarial attacks, to using domain knowledge or other forms of auxiliary information to improve the robustness of predictions. Additionally, researchers are also exploring the use of formal methods and reasoning techniques to prove that models are robust against a wide range of potential attacks, and to identify potential weaknesses and vulnerabilities in advance. By addressing these challenges and developing effective countermeasures, researchers and practitioners aim to ensure that AI and machine learning systems can be deployed securely and with confidence, and that the benefits of these technologies can be fully realised \cite{ilahi2021}.

The first reported introduction of adversarial attacks in neural networks was by Szegedy et al. \cite{szegedy2013}. However, the lack of understanding of these attacks and the newness of the subject area makes it difficult to criticise the potential limitations of the methods suggested for defence. A taxonomy of adversarial attacks has been developed through specific criteria and threat models, as proposed by McDaniel et al. \cite{mcdaniel2016}. The defence against adversarial attacks can be classified into supervised and unsupervised techniques. Supervised strategies aim to improve the generalisation of the learning models and to tailor specific perturbation patterns. Unsupervised strategies aim to smooth the decision boundaries through regularisation of loss functions and the removal of nuisance variables. The robustness of the decision boundary is crucial in the defence against adversarial attacks and requires more attention. The historical timeline of adversarial machine learning and the current limitations in secure machine learning algorithms for detecting malware have been discussed by Biggio and Roli \cite{demontis2019}. 

The use of adversarial training was first introduced in \glspl*{dnn} to make the functions resistant to adversarial perturbations and maintain the accuracy of clean inputs. According to Goodfellow et al. \cite{goodfellow2014}, this was achieved by relying on the linearisation of lost data points. However, robust models against adversaries can still be vulnerable to more iterative attacks. The ImageNet recently observed the capacity for adversarial training using step methods. Training against multi-step methods is expected to have better resistance against adversaries, as indicated by Madry et al. \cite{madry2017}. Although the adversarial inputs may look similar to the original inputs, the linearity of \glspl*{dnn} makes it possible for malicious perturbations to impact decision-making. It is vital to note that perturbations can only address certain types of attacks, as pointed out by Carrara et al. \cite{carrara2019}.


Adversarial attacks refer to inputs that aim to exploit flaws in a detection system and manipulate the behaviour of trained models. Adversarial mitigation aims to counter these attacks by making the models more robust to malicious inputs, both during the training and decision-making phases. Attackers can use two different techniques: data poisoning, which involves altering a few training inputs to mislead the model's output, and model poisoning, in which the attacker tries to direct the model to produce a false label for perturbed instances \cite{catak2020}. The issue of poison target attacks, used against \glspl*{dnn}, is widespread and exploits the influences of the functions. Koh and Liang \cite{koh2017} developed training with small perturbations to specific training points, making predictions for the target set test points. \glspl*{dnn} are altered by adding perturbations to input vectors, limited to modifying only one pixel in an image. For image recognition, the accuracy is maximised. The \gls*{dnn} adds noise and changes one pixel to alter the \gls*{cv}, which is represented by the probability distribution used by the \gls*{de} algorithm to filter pixels and determine the correct image class. The fast-gradient sign algorithm is primarily used to calculate the effectiveness of perturbations for the hypothesis \cite{su2019}. The linearity and high dimensionality of inputs lead to sensitivity to perturbations. This requires consideration for image classification, as it only works for small size (e.g. $32\times 32$) images by modifying a specific colour to lower the network's confidence in iterations and resulting in new categories with the highest classification confidence being successfully chosen \cite{su2019}.

Adversarial attacks pose a threat to the accuracy and reliability of machine learning models. These attacks use bi-level optimisation control, making them difficult to detect. Mitigation strategies to defend against these attacks include shielding adversarial inputs through normalising raw inputs, rate limiting, regression testing, anomaly detection, and human intervention. Rate limiting involves limiting the number of users who can contribute to the model and using mechanisms to prevent false positives and negatives \cite{munoz2019}. Regression testing involves comparing a newly trained model to baseline standards and estimating changes in outputs of previous models. Anomaly detection locates suspicious patterns through metadata, IP-based analysis, heuristic analysis, analysis of temporal dynamics, and graph-based Sybil attack detection methods\cite{munoz2019}. Machine learning models can be attacked in various forms such as spam emails, phishing attacks, and malware. The main types of adversarial attacks are white box and black box attacks \cite{catak2020}. In a white box attack, the attacker has complete access to the target model's network and can exploit its vulnerabilities. In a black box attack, the attacker does not have complete access but can still affect the model's performance.Another type of adversarial attack is the grey box attack which combines elements of both white box and black box attacks \cite{catak2020,wang2020}. Adversarial perturbation takes advantage of the geometric correlation between decision boundaries in classifiers. To counter adversarial attacks, methods have been proposed such as modifying the training process, changing the network function, and using external models for classification. A taxonomy has been created to categorise these defence methods based on whether they modify the original network or add to its architecture. The effectiveness of these defences against different attack methods is yet to be fully determined, and certification is required to ensure their reliability  \cite{akhtar2018}.

\glspl*{gan} are a type of generative model that can produce new content based on a dataset. The unsupervised learning method can analyse, capture, and copy variations in the data \cite{yang2018}. The model can add noise to improve its confidence in predictions, but this can also result in misclassifications. The generator and discriminator in a \gls*{gan} can be used to generate fake samples of data and detect deepfakes. Face swapping was initially used for entertainment purposes on social media, but advancements in deep learning have made it more realistic and difficult for forensics models to detect the fakes \cite{korshunova2017}. Deepfakes use superimposition of faces to create synthetic media and have gained popularity due to the improved realism produced by the application of auto-encoders and \glspl*{gan} in facial recognition and media tracking. Deepfakes were first introduced in 2017 for creating synthetic media using face swaps from adult entertainment but have since had negative consequences for political campaigns due to falsification \cite{badrinarayanan2017, zamyatin2018}. While deepfakes have positive applications, such as being used in movies to reshoot scenes, the negative consequences and potential harm outweigh the positive implications \cite{demetrious2021}.

Traditional methods for detecting deepfakes in images were introduced using CNN models and face-cropping techniques. Dolhansky et al. \cite{dolhansky2019} used the \gls*{cnn} model with six convolution layers and TamperNet to detect low image manipulation \cite{skibba2020}. Later, XceptionNet was introduced and trained on full-sized images using separable convolution layers, which improved the precision of TamperNet from 79.0\% to 83.3\% and the recall from 3.3\% to 26.4\% \cite{wang2018}. Deepfakes detection currently involves face-tracking and a \gls*{cnn} classifier to determine if an image is real or fake. However, the current methods are vulnerable to adversarial attacks, which can mislead the classification, and a robust deepfake detector is still needed. The state-of-the-art method, iCaps-Dfake, uses different nature feature extraction to reduce the need for pre-processing, but it fails to consider the potential for human-created fake images \cite{khalil2021}. The high potential risk of misinformation and disinformation posed by deepfakes is a growing threat to cybersecurity, especially in areas such as non-consensual pornography, political disinformation, and financial fraud. However, full implementation of deepfake detection methods would require large sample datasets, which can be difficult to access.

\section{Methodology}\label{sec:method}
The application was implemented in Python 3+, using \gls*{opencv} for face recognition, NumPy/SciPy for computational linear algebra, and PyTorch/TensorFlow for \gls*{ml} and \gls*{dl} techniques. Jupyter Notebook was used as a web-based interactive computational environment for larger datasets for data analysis, data visualisations, and exploratory computing. OpenCV was incorporated for pre-processing the dataset to handle this data type. The project mainly used TensorFlow for high-level model development with more mature libraries and built-in artificial intelligence. The framework is prevalent under \gls*{cpu}, but for faster training potential, \gls*{gpu} such as \gls*{cuda} can be used. Nevertheless, the machine used for the project was an Intel(R) Core™ i7 \gls*{cpu} @ 1.80GHz with 64GB of \gls*{ram} and was not equipped with an NVIDIA \gls*{gpu}. 

The project primarily used two datasets: the \gls*{dfdc}\footnote{Available online, \protect\url{https://www.kaggle.com/c/deepfake-detection-challenge}, last accessed: 15/02/2023} and COVID\footnote{Available online, \protect\url{https://www.kaggle.com/datasets/pranavraikokte/COVID19-image-dataset}, last accessed: 19/02/2023}. These datasets were chosen due to their ability to be used on wide range of models and techniques. The criteria for selecting these datasets included two classes, a large number of images or videos, and little to no noise in the real class. Noise in the dataset would include low resolution, poor quality images with low pixel count and irrelevant images. A large number of visual representations in each class is crucial to detecting deepfakes by adopting training within Python. The \gls*{cnn} requires a large number of images to differentiate between the classes. To create the fake visuals, the project uses competing neural networks between the generator network and the discriminator network. The noise in the dataset needs to be minimal for the optimum processing of data, as generating poor results would not apply to the class.

Furthermore, the Real vs. Fake dataset is a large collection of non-celebrity face images, with 140,000 faces equally split between real and fake, collected by NVIDIA. These images were generated using StyleGAN and have been standardised by resizing to 256 pixels in JPG format. The dataset is split into train, test, and validation sets, with a 60:20:20 ratio, and each set has a corresponding CSV file. The dataset contains a variety of genders, ages, races, and facial features. Although unsupervised clustering was suggested for the dataset by the \gls*{dfdc} Kaggle forum, it was considered out of the scope of this project. The test folder contains unclassified images that participants can use to test the accuracy of the models they create. The \gls*{dfdc} dataset \cite{dfdc2020} was released by Facebook and consists of 50 files of training data, with a total size of 470 GB, and a metadata JSON file with labels. The dataset is split into train and test sets in an 80:20 ratio. The test set contains a smaller sample of MP4s, which is used for validation. The videos were recorded under various conditions, with different lighting and backgrounds and a variety of people. We also tested the models against the COVID dataset  which is composed of  137 cleaned images of COVID-19 based images and more 317 in total containing Viral Pneumonia and Normal Chest X-Rays structured into the test and train directories.

\begin{figure}[]
	\centering
	\includegraphics[scale=0.6]{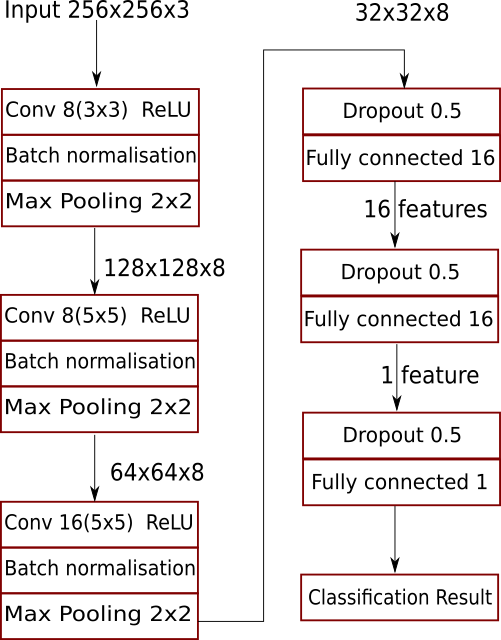}
	\caption{Customised 7-layer \gls*{cnn} architecture. The network is composed of 3 convolutional layers, 3 fully connected layer and the classification layer.}
	\label{fig:cnn}
\end{figure}

The following step was to train and test the deepfake detection framework using \gls*{cnn} (see Figure~\ref{fig:cnn}) and \gls*{gan}. Before training the model, the Real vs. Fake dataset was pre-processed, and the images resized to $256\times 256$ JPG format. The dataset was split into 60\% for training, 20\% for validation, and 20\% for testing, which allows the model to learn patterns to classify faces in the training set and validate the learning on the validation set before testing on the test set. This approach is important for preventing overfitting, which occurs when the model performs well on the training data but is unable to generalise to new data. While other datasets only have a train and test set, the Real vs. Fake dataset has a validation set to help the model generalise better. To ensure the model is effective, it was important to use a large sample size in each class while avoiding excessive data that could be time-consuming to process under \gls*{cpu}. The model's ability to distinguish between the labelled sets will be tested using the test set. It was crucial to accurately classify the detection to avoid potential errors caused by overfitting or underfitting.

To customise the \gls*{cnn} and \gls*{gan} models, the pre-trained models were split into specific algorithms with distinct uses. CNN, for example, was used for two-dimensional image classification and can learn position and scale variants in the data structure using spatial relationships. ResNet50 was used to boost neural network performance by adding more layers to the network and finding the predicted and residual of each result. \gls*{gan} was used to distribute any data for generating instances, which is specifically useful for deepfake detection. To evaluate the performance of the models, appropriate parameters such as epochs, batch size, and activation filters were standardised to enable comparisons and detect issues.

During the testing phase, it is possible that the custom model may produce higher error rates due to false negatives that occur in the computational environment compared to real-world scenarios. This is a consequence of not having optimal labels with or without parallax errors that affect the position of the image or video. To ensure accurate results, it was important to clean the dataset of all noise and reduce the margin for error. The \gls*{dfdc} video dataset presents a different challenge as it is in MP4 format and requires analysis frame-by-frame for detection. The \gls*{cnn} classifies each frame as real or fake, and the goal is to minimise the adversary perturbation from the delta that was used to create the adversarial input, which is discussed further below. To classify an image in PyTorch, the first step is to normalise it by subtracting the mean and dividing the result by the standard deviation. This transformation is applied using PyTorch layers, and the resulting image is resized to $224 \times 224$ pixels, which is the standard input size. A pre-trained ResNet50 model can then be used to classify the image, producing a probability vector by applying a SoftMax operator. The classifier can be fine-tuned to adjust the probabilities based on the specific task at hand. This is done using the hypothesis function:

\begin{equation}
h_{\theta}:\chi \to \mathbb{R}^{k}
\end{equation}

where $h_{\theta}$ is the mapping between input space and output space, which is a k-dimensional vector. Here, k is the number of classes predicted; $\theta$ represents all parameters within the model (such as convolution filters or fully convolved layers), which are typically optimised while training the neural network. The precision of the model object is linked to $h_{\theta}$. The loss function is given by:

\begin{equation}
l:\mathbb{R}^{k}.\mathbb{Z}{+}\to \mathbb{R}{+}
\end{equation}

Here, $l$ maps the model's prediction and labels of non-negative numbers. The model output for the logics can be positive and negative for the second argument in the index for distinguishing the true class. For $x \in X$ the input and $y \in \mathbb{Z}$ the true class, the loss function computes the difference between the model's prediction and the true label. The most prevalent form of loss in deep learning is the cross-entropy loss (SoftMax), defined as:

\begin{equation}
l(h_{\theta}(x),y)=log\left ( \sum_{j=1}^{k}exp(h_{\theta}(x_{j})) \right )-h_{\theta}(x)y
\end{equation}

Here, $h_{\theta}(x_{j})$ denotes the $j^{th}$ term for the vector $h_{\theta}(x)$. The adversarial example is used to manipulate the loaded image, which enables the classifier to lower the probability of recognising the deepfake image. The approach is used to train for optimisation on the parameters to minimise the average loss in the training set:

\begin{equation}
min=\frac{1}{m}\sum_{m}^{i=1}l(h_{\theta}(x_{i}),y_{i})
\end{equation}

Here, the parameter $\theta$ is optimised to minimise the average loss over part of the training set. Another form involves adjusting the image to optimise the network parameters that will adjust the image for maximising the loss. This is done to solve the optimisation problem with the equation below:

\begin{equation}
max=l(h_{\theta}(\hat{x}),y)
\end{equation}

Here, $\hat{x}$ denotes the adversarial example for maximising the loss. The gradient step is used for adjusting the optimiser to perturbations for maximising the loss for the training set. This allows computing the probability and then adding the delta as random noise for the adversarial attack, which creates an image that looks identical, but the probability of classifying it is significantly reduced. Finally, targeted attacks can also be used to maximise the loss of a specific target class in the probability vector, by minimising the loss of that target class through linear optimisation. Adversarial robustness is a significant challenge when it comes to attacking deep learning classifiers, as these attacks aim to modify existing images in the training set. It is important to consider the potential adversarial risks that may arise, such as retaining specific regions of perturbations to encode the images.

\section{Results analysis}\label{sec:results}
Organising the dataset into the correct format is critical for the success of the project. One of the main challenges we encountered was resizing the images or video frames to dimensions suitable for specific models. Another issue was the limited availability of real images of target faces, which made it difficult to create a balanced dataset for binary classification. In comparison, there were more deepfake faces available in the real-life dataset. We started with binary classification, a supervised machine learning technique that involves making observations and classifications. We used a simple \gls*{cnn} to train the custom model on a variety of different faces in the Real vs. Fake dataset, and then moved on to the \gls*{dfdc} dataset. Since we were learning a new language, training the model on a \gls*{cpu} was time-consuming. To generate predictions, we trained a single model with balanced classes, and the dataset was categorised into distinct classes for labelling and targeting purposes. We also found that we needed to convert the \gls*{dfdc} dataset into images to apply any algorithm once frame-by-frame detection had occurred.

In deep learning, the mapping process is used to detect deepfakes by categorising them as either real or fake. \glspl*{cnn} are particularly effective in image recognition and are commonly used to train the dataset. The process involves training the weights with input images and validating the performance of the network. The input images are processed through a series of layers, including convolution layers, pooling layers, and fully connected layers, which are then activated by a sigmoid function. This efficient image recognition and classification is what makes \glspl*{cnn} an ideal technique for detecting deepfakes. This process can be visualised through the mathematical equation:

\begin{equation}
n(out)=\left [ \frac{n(in)+2p-k}{s} \right ]+1
\end{equation}

where $n(in)$ is the number of input features, $n(out)$ is the number of output features, $k$ is the convolution kernel size, $p$ is the convolution padding size, and $s$ is the convolution stride size.

\begin{figure}[]
	\centering
	\includegraphics[scale=0.4]{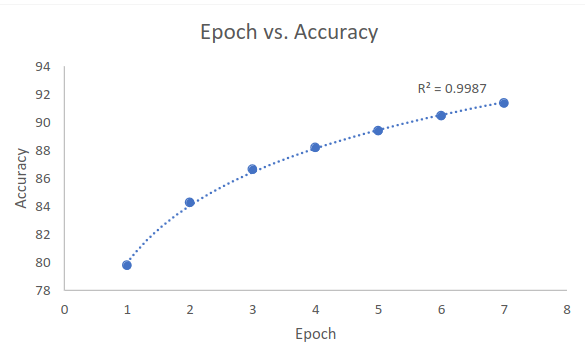}
	\caption{The epochs and Accuracy on Real vs. Fake dataset with \gls*{cnn} model}
	\label{fig:res2}
 \vspace*{-5mm}
\end{figure}

The model summary provides insight into the architecture of the model, which includes 6 layers consisting of 4 convolution layers and 2 fully connected layers. To identify the 80,000 images belonging to 2 classes - real (1) or fake (0) - binary encoding is applied. The generator is then initialised to flow the data to the fit generator for training the model and validating its accuracy. As shown in Figure \ref{fig:res2}, the blue line represents the validation loss, while the orange line represents the validation accuracy. The trained CNN model demonstrated a precision value of 0.748, which indicates that around three quarters of the images were classified correctly. It's expected that larger models, such as ResNet50, may produce better precision, as their features are more selective. The number of epochs refers to the full pass of the training set through an algorithm, and hyperparameters can be specified to determine the number of epochs. The model's internal parameters are updated with each epoch, which results in the gradient learning algorithm. The epoch number and factor rate will become zero over time as the algorithm learns. Achievable reasonable test correctness is required to apply this complexity to real-world applications. To achieve a less biased estimate and enhance the model, k-fold cross-validation can be used. 


When evaluating classification performance, accuracy is a common metric used to measure how well the model has correctly classified the samples. It can be calculated using the equation:

\begin{equation}
Acc = \frac{TP + TN}{TP + FP + TN + FN}
\end{equation}

where $TP$ is the number of true positives, $TN$ is the number of true negatives, $FP$ is the number of false positives, and $FN$ is the number of false negatives. The accuracy score gives the proportion of correctly classified samples out of the total number of samples. 

Accuracy is a common metric used to assess the performance of a classification model. It is calculated as the ratio of correctly classified samples to the total number of samples. However, accuracy alone does not take into account the loss function used by the model. To address this issue, the cross-entropy function is often used instead of accuracy. A high accuracy value is desirable, but the focus should also be on maximising the cross-entropy function to achieve the best results. Figure \ref{fig:res2} shows the results of using the cross-entropy function to measure the performance of a model. The R2 value of 0.9987 indicates high accuracy of the predictions, but increasing the number of epochs may further improve the results. When training the model, it was found that the precision increased as the number of epochs increased. This suggests that the model is learning the specific patterns that are important for detecting deepfakes. In comparison to a custom model, the pretrained convolutional base used in this study achieved higher accuracy with consistent performance across different epochs. The pretrained model is tuned for quicker training time and is able to classify real and deepfake images with high accuracy. The study's accuracy of 87.3\% for detecting deepfakes is higher than that reported in the literature for the \gls*{dfdc} dataset (79\%), which could be attributed to the larger Real vs. Fake dataset used in this study. Overall, the results suggest that using a pretrained model can lead to better performance for deepfake detection.

The classification model's matrix parameters are set to predict the majority class frequency for null error rate conditions. This means that the model's error rate is preferred to be higher than the null Real error rate. The \gls*{roc} curve is used to evaluate the classifier's performance, and a desirable threshold is chosen, as seen in Figure \ref{fig:res2}.

\begin{figure}[]
	\centering
	\includegraphics[scale=1.8]{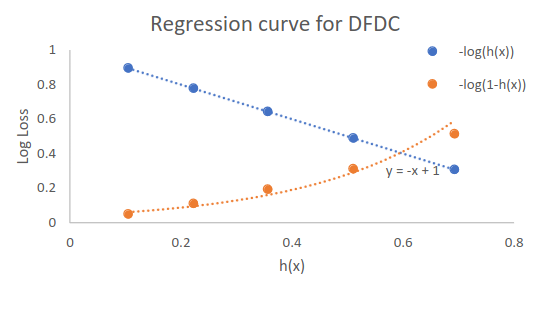}
	\caption{Logarithmic loss using logistic regression for \gls*{dfdc} dataset for inception at 0.6}
	\label{fig:res3}
 \vspace*{-5mm}
\end{figure}

The linear regression model with the same model and parameters can also be used for classification, but if the classification exceeds 1, logistic regression is necessary to predict real and fake labels. The error function of the linear regression model includes weighting parameters that are used in the gradient descent optimisation algorithm to minimise the misclassification error. Figure \ref{fig:res3} shows the logarithmic loss metric, which evaluates the probability of correctly classifying the data. This metric can be used to compare models based on the following equation, primarily for binary classification:

\begin{equation}\label{eq:log}
-\frac{1}{N}\sum_{i=1}^{N}y_{i}.log(p(y_{i})+(1+y_{i}).log(1-p(y_{i}))
\end{equation}

In this equation, the first term is zero for the binary cross-entropy model when the actual class is 1 (real) and the other actual class is 0 (fake). Additionally, the binary cross-entropy model is less exponential than typical regression curves and tends to be more symmetric. This could be due to the possibility of overfitting when there is no validation set, as shown in Figure \ref{fig:res3}.

Equation \ref{eq:log} shows the binary cross-entropy model used primarily for binary classification. When the actual class is 1 (real), and the other actual class is 0 (fake), the first term of the equation becomes 0. Unlike typical regression curves, this model is seen to be less exponential and tends to hold more symmetry. However, this could be due to the potential of overfitting, as shown in Figure \ref{fig:res3}, which may result from the absence of a validation set. Most models would have concluded that the entire video is real, regardless of whether a single frame is fake. However, state-of-the-art detection confirms that each frame needs to be investigated. The frame-by-frame detection can later be used in a white-box setting to propagate the loss in the entire model when adding perturbations, obtaining a gradient for adversarial frames. This is because current deepfake detection DNNs are assumed to be non-adaptive in attacks that aim to withstand human eyes, generating realistic fake videos.

\begin{figure}[]
	\centering
	\includegraphics[scale=1.8]{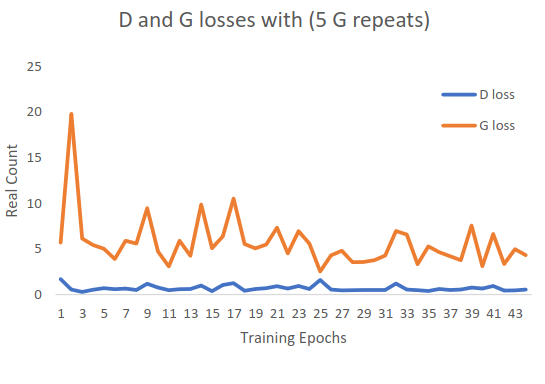}
	\caption{\gls*{gan} graph for D and G losses in Real vs. Fake dataset strength of real estimates for the epoch trained. The Y axis is the real indices that has been standardised and the x-axis represents the epoch}
	\label{fig:res4}
 \vspace*{-5mm}
\end{figure}

The customised \gls*{gan} used to implement the losses of both the discriminator and generator models, and their training process does not exhibit a clear pattern. Recent advancements in deep learning have leveraged data augmentation techniques, such as cropping, flipping, and zooming, to improve model performance in a given domain, typically applied to image data. In \glspl*{gan}, the generator and discriminator compete against each other, with each trying to achieve a higher loss than the other, creating an environment for learning from the training data based on the received loss. In Figure \ref{fig:res4}, the convergence of the Real vs.  Fake plot indicates that the model has reached a maximum level of performance, and any additional enhancements are unlikely. It is worth noting that the D loss is less interactive than the G loss, which is not counterbalanced as in a traditional model. This implies that the generator will be improved in the next iteration, and synthetic observations of good quality can be generated. A generator with 100\% accuracy would generate synthetic samples that the discriminator could classify as real, whereas a discriminator with 50\% accuracy would be unable to distinguish fake observations from real ones.

\begin{figure}[]
	\centering
	\includegraphics[scale=1.8]{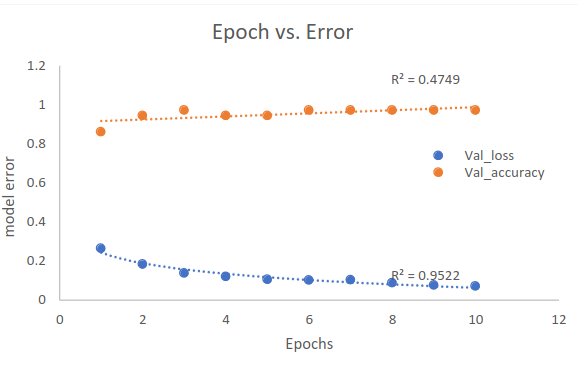}
	\caption{The epochs and validation for COVID dataset against validation loss (blue) and validation accuracy (orange) with \gls*{cnn} model.}
	\label{fig:res5}
 \vspace*{-5mm}
\end{figure}

To illustrate how these models can be applied to real-world scenarios, let's consider the example of a COVID dataset. As shown in Figure \ref{fig:res5}, this dataset was used to train a binary classification model with a limited amount of data. Since there were only 100 images in each class, the model had to be carefully adapted to these weights in order to learn to accurately classify COVID positive and negative cases. It's worth noting that in real-life scenarios, obtaining a large, diverse dataset can be challenging, and may require a significant amount of effort to collect and label the data. Nonetheless, even with limited data, it's possible to train a model that can provide valuable insights and predictions. As the field of deep learning continues to evolve, new techniques and approaches are being developed to address these challenges and improve the performance of models with limited data. 

White box perturbations refer to a type of attack where an adversary has access to a machine learning model's architecture and parameters. By exploiting this information, the adversary can create an output gradient that generates an adversarial example using the model's gradient and loss. This is done by adding perturbations to the input, which can cause the model to incorrectly classify the input, moving it closer to the decision boundary between the true class (real) and the other class (fake). The goal of white box perturbations is to decrease the predictability of the input and generate an attack that is imperceptible to the human eye.

In the image domain, the Lp norms are used to measure the amount of perturbation by quantifying the maximum distortion added to individual pixels in the image. The primary objective of adversarial perturbations is to generate an attack that is imperceptible to the human eye. For example, if a model's initial prediction for the true class probability of a given image is 10.9, indicating that it is a real image, an adversarial attack can reduce the true class probability to $2.26e^{-18}$ by subjecting the image to a delta (perturbation), leading to misclassification.

Table \ref{tab:res1} shows the Probability of true class in Real vs. Fake dataset as the iterations of each degradation that the attack applies to the image, ultimately resulting in misclassification. In summary, white box perturbations allow adversaries to create adversarial examples by exploiting a model's parameters and architecture. These attacks are difficult to detect because they generate imperceptible changes to the input, making them a serious threat to the security of machine learning models.
\begin{table}[]\caption{Probability of true class in Real vs. Fake dataset before adding delta for perturbation.}\label{tab:res1}
\vspace*{-5mm}
\begin{center}
\begin{tabular}{|c|c|} 
\hline
Delta & True Class probability \\ \hline
0     & 10.90469               \\ \hline
5     & 35.33980               \\ \hline
10    & 40.24509               \\ \hline
15    & 37.24498               \\ \hline
20    & 37.62185               \\ \hline
25    & 40.85791               \\ \hline
\end{tabular}
\end{center}
\vspace*{-5mm}
\end{table}

It is important to note that the adversarial attack significantly degrades the accuracy of the model, reducing the probability of correctly detecting an actual attack. While the cost of an error on the adversarial input may not be high in real-life applications, a decrease in accuracy can have severe consequences. In particular, classification errors can lead to the abstention of detecting inputs, and the significant amount of noise applied in these datasets can result in further classification refusals. It's crucial to consider the domain of the system being used, as the difference in error rates between random inputs and adversarial inputs can vary depending on the application.

To ensure the validity of the testing results, it was crucial to avoid using the adversarial attacks on the training set, which is a common pitfall in evaluating the robustness of models. This was especially important when applying the attacks to a defence mechanism, as the overfitting can be against a particular attack that was included in the training set. Some attacks may be effective on both training and testing sets, which could potentially overestimate the model's robustness. Therefore, the strongest attack against a custom model requires a specifically tuned approach tailored to the model's architecture. To evaluate the potential of invalidating the model's robustness, it was necessary to consider errors that may deviate from classifying the correct class. To ensure the testing is effective, it was essential to define a precise threat model that assumes the attacker's knowledge, goals, and capabilities. Finally, to perform an adaptive upper bound of robustness, the loss function should be changed to cause misclassification, while the defence mechanism is evaluated.
\section{Conclusions and Future Work}\label{sec:conclusions}
This article proposes a state-of-the-art DeepFake detector network that utilises machine learning techniques to mitigate adversarial attacks on autonomous systems that are vulnerable to deepfakes. To evaluate the efficacy of the proposed approach, a transferable white-box attack via perturbations was developed to pose a practical threat to the deepfake detection system. The attack was applied to a variety of \gls*{dfdc} and COVID datasets, to demonstrate the effectiveness of the proposed algorithms on \gls*{dl} techniques such as \gls*{cnn} and \gls*{gan}. The proposed approach showed promising results for detecting deepfakes, achieving a precision value of accuracy of 76.2\% on the \gls*{dfdc} dataset. However, this result can be improved by increasing the volume of media in the datasets. The detection models were able to distinguish between real and fake media and can be applied in real-life scenarios such as the COVID dataset. To evaluate the system's robustness against adversarial attacks, white-box attacks via perturbation were applied to distort the images. The Real vs. Fake dataset's classification probability was reduced from 10.9 to $2.26e^{-18}$ with the addition of perturbation, and the COVID dataset's probability was reduced from 11.1 to $1.46e^{-16}$. The proposed method offers an effective approach to detecting deepfakes by incorporating adversarial examples through white-box attacks that use perturbations to decrease classification probability.

The proposed work on mitigating adversarial attacks and detecting deepfakes has significant ethical impacts. Deepfakes, which are manipulated media that appear real, can be exploited to deceive, manipulate public opinion, spread misinformation, and facilitate cybercrime. The proposed DeepFake detector network aims to address these ethical concerns by safeguarding truth and trust in digital media, mitigating social and political manipulation, and protecting privacy and consent. This work has implications for individuals, organizations, and society as a whole.The proposed work on mitigating adversarial attacks and detecting deepfakes has significant ethical impacts. Deepfakes, which are manipulated media that appear real, can be exploited to deceive, manipulate public opinion, spread misinformation, and facilitate cybercrime. The proposed DeepFake detector network aims to address these ethical concerns by safeguarding truth and trust in digital media, mitigating social and political manipulation, and protecting privacy and consent. This work has implications for individuals, organizations, and society as a whole. From a business perspective, the impact of \gls*{ai} in relation to deepfake detection and \gls*{ai}-based technologies is notable. By incorporating \gls*{ai}-powered deepfake detection systems, businesses can enhance trust and reliability in their digital media content. This helps demonstrate their commitment to authentic communication, transparent marketing practices, and reliable representation of products or services. Additionally, implementing robust deepfake detection mechanisms safeguards brand reputation by proactively identifying and addressing potential threats that manipulate brand images or spread false information. Furthermore, integrating deepfake detection systems into cybersecurity strategies strengthens defences against malicious activities such as impersonation, unauthorised access, and fraud, ensuring the integrity and authenticity of digital interactions. From a business perspective, the impact of \gls*{ai} in relation to deepfake detection and \gls*{ai}-based technologies is notable. By incorporating \gls*{ai}-powered deepfake detection systems, businesses can enhance trust and reliability in their digital media content. This helps demonstrate their commitment to authentic communication, transparent marketing practices, and reliable representation of products or services. Additionally, implementing robust deepfake detection mechanisms safeguards brand reputation by proactively identifying and addressing potential threats that manipulate brand images or spread false information. Furthermore, integrating deepfake detection systems into cybersecurity strategies strengthens defences against malicious activities such as impersonation, unauthorized access, and fraud, ensuring the integrity and authenticity of digital interactions.

In terms of future work, this study can serve as a starting point for generating perturbations by minimising the likelihood of correct class classification. Another objective could be to work on detecting deepfakes in audio for video media. Additionally, further advancements in the field of deepfakes would require obtaining larger datasets and training models using \gls*{opencv} architecture in Python. To handle increased input size, it would be necessary to standardise parameters such as image or video size. This can help improve the model's performance. In addition, a promising direction of research would be to work on a physical "adversarial patch" which could enhance facial recognition by allowing the classifier to add any image or video to the chosen target class, misclassifying the arbitrary modification of each pixel, ultimately defending against adversarial attacks. Lastly, the methodology developed in this study could be extended beyond the forensic domain of deepfake detection, to enable 3D object localisation for use in robotic applications such as self-driving vehicles.

\bibliographystyle{IEEEtran}
\bibliography{IEEEabrv,references}

\end{document}